\documentclass[final]{cvpr}
\usepackage{times}
\usepackage{epsfig}
\usepackage{graphicx}
\usepackage{amsmath}
\usepackage{amssymb}
\usepackage{multirow} 
\usepackage[ruled]{algorithm2e}
\usepackage{subfigure}
\usepackage{booktabs}
\usepackage{bm}

\usepackage{floatrow}
\newfloatcommand{capbtabbox}{table}[][\FBwidth]

\usepackage{makecell}
\usepackage{algorithmic}

\usepackage{color}
\definecolor{blue}{rgb}{0, 0, 1}

\usepackage[pagebackref=true,breaklinks=true,colorlinks,bookmarks=false]{hyperref}

\def\cvprPaperID{5906} 
\def\confYear{CVPR 2021}

\begin{document}

\title{Look Before You Leap: Learning Landmark Features \\ for One-Stage Visual Grounding}

\author{

Binbin Huang \textsuperscript{\rm 1}
\quad
Dongze Lian \textsuperscript{\rm 1} 
\quad
Weixin Luo \textsuperscript{\rm 1} 
\quad
Shenghua Gao\textsuperscript{$\dag$}\textsuperscript{\rm 1,2}
\and 
\textsuperscript{\rm 1}ShanghaiTech University 
\\
\textsuperscript{\rm 2}Shanghai Engineering Research Center of Intelligent Vision and Imaging \\
{\tt\small \{huangbb, liandz, luowx, gaoshh\}@shanghaitech.edu.cn}
}

\newcommand\blfootnote[1]{%
 \begingroup
 \renewcommand\thefootnote{}\footnote{#1}%
 \addtocounter{footnote}{-1}%
 \endgroup
}

\maketitle

\begin{abstract}

An LBYL (`Look Before You Leap') Network is proposed for end-to-end trainable one-stage visual grounding. 
The idea behind LBYL-Net is intuitive and straightforward: we follow a language's description to localize the target object based on its relative spatial relation to `Landmarks', which is characterized by some spatial positional words and some descriptive words about the object. The core of our LBYL-Net is a landmark feature convolution module that transmits the visual features with the guidance of linguistic description along with different directions. Consequently, such a module encodes the relative spatial positional relations between the current object and its context. Then we combine the contextual information from the landmark feature convolution module with the target's visual features for grounding. To make this landmark feature convolution light-weight, we introduce a dynamic programming algorithm (termed dynamic max pooling) with low complexity to extract the landmark feature. Thanks to the landmark feature convolution module, we mimic the human behavior of `Look Before You Leap' to design an LBYL-Net, which takes full consideration of contextual information. Extensive experiments show our method's effectiveness in four grounding datasets. 
Specifically, our LBYL-Net outperforms all state-of-the-art two-stage and one-stage methods on ReferitGame. On RefCOCO and RefCOCO+, Our LBYL-Net also achieves comparable results or even better results than existing one-stage methods. Code is available at {\small \url{https://github.com/svip-lab/LBYLNet}}.
\end{abstract}
\blfootnote{$\dag$~Corresponding author}

\begin{figure}
    \centering
    \includegraphics[width=\linewidth]{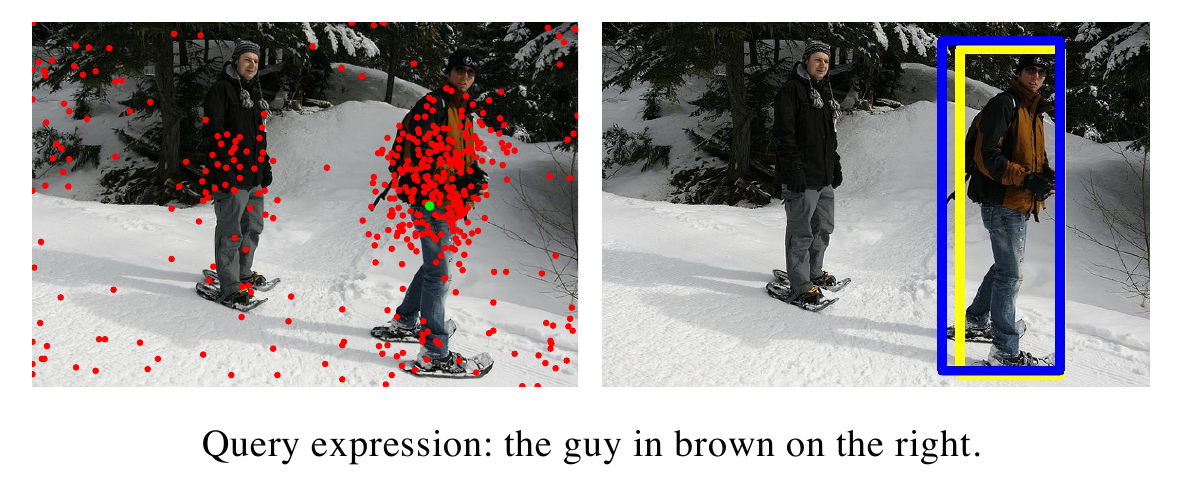}
    \caption{ Illustration on how LBYL-Net uses contextual cues. On the left figure, the target location (green) perceives information from landmarks (red) to localize itself. In this case, landmarks attend to the attribute \textit{brown} to differ from the other \textit{guy}. The right figure shows our predicted result (blue box) and the ground-truth (yellow box).}
    \label{fig:my_label}
\end{figure}

\section{Introduction}

Humans often refer to objects in an image by describing their relationships with other entities, \emph{e.g.} ``laptop on table'', and understanding their relationships is vital to comprehend referential expressions. \textit{Visual grounding}, aiming to localize the entities described by referential expressions, inherently requires contextual information for grounding the target. By considering relationship of objects, several recent studies have achieved promising results \cite{yu2016modeling,nagaraja2016modeling,Hu2017,yu2018mattnet}. In particular, these methods usually leverage a two-stage paradigm, where they first extract region proposals as candidates and then rank the region-expression pairs as a way of \textit{metric learning}. 

Although effective, these two-stage methods have the following defects: (i) two stages bring time complexity, which hinders these methods from being real-time. (ii) since only objects in the pre-defined categories are considered, the contextual cues in the whole scene may not be fully exploited. Motivated by the success of one-stage detection \cite{Redmon2018,Liu2016}, one-stage based visual grounding has gained great interest, where the pipeline is simplified, and the inference is accelerated with a \textit{detecting and matching simultaneously} paradigm \cite{yang2019fast,sadhu2019zero}. These detection-based one-stage approaches, however, still perform localization on grid features \textit{indivisually}. The contextual information in the whole scene, especially relationships between objects, is not thoroughly investigated yet, making them inferior to their two-stage counterparts. 

From this perspective, it is desirable to enable relationship modeling in one-stage visual grounding since the object requires perceiving the relational entities mentioned by the language to localize itself, \emph{e.g.} ``the chair with owl on it''. 
We enable the grid features to capture rich contextual cues for better localization by introducing the concepts of \textit{Landmark Features} and \textit{Landmark Feature Convolution}. 

To begin with, in our real life, we usually judge our location or positions of other buildings by using an easily noticed building, which is called \textit{Landmark}. Similarly, in the image domain of visual grounding, the landmarks can be regarded as those locations that are helpful for object localization. Figure \ref{fig:my_label} shows the visualization of landmarks in an image given the query language. These landmarks might fall on the background, other objects or the object itself to be located as long as they have helpful semantic cues. The network could extract the \textit{Landmark Features}, which contains the global contextual information from these landmarks. To fully integrate this contextual information to improve the localization, these landmark features are propagated to the target object from different orientations to characterize relative positions by an efficient dynamic programming algorithm, termed \textit{Dynamic Max Pooling}. By aggregating landmark features with a standard convolution operation, the grid features are equipped with (i) global receptive field (ii) direction-awareness. We call the whole process as \textit{Landmark Feature Convolution}.

Considering the long-range context, we propose a novel one-stage visual grounding framework. Our network first applies feature pyramid network (FPN) \cite{lin2017feature} to extract visual features of objects from different scales, of which effectiveness has been proven for better object localization. A landmark feature convolution is then employed to extract contextual information of objects from different orientations, for a better characterizing relationship to objects mentioned by expressions. Since we mimic the `Look Before You Leap' behavior of us humans in visual grounding, we term our method as LBYL-Net.

We summarize our main contributions as follows:
\begin{itemize}
	\item We propose a novel LBYL-Net for one-stage visual grounding, which combines the visual feature of objects mentioned in the description as well as landmark features of the spatial relationships between different objects for target localization; 

	\item A landmark features convolution is proposed, which has a global receptive field but without introducing extra parameter and complexity. We showcase it's superiority over related convolutional modules, \emph{i.e.} dilated convolution \cite{yu2015multi}, deformable convolution \cite{dai2017deformable} and Non-Local module \cite{wang2018non}.
	\item Extensive experiments show the effectiveness and efficiency of our LBYL-Net on four grounding datasets. Especially, our method achieves state-of-the-art performance on ReferitGame. 

\end{itemize}

\section{Related Work}
\noindent \textbf{Two-Stage Visual Grounding.}
Probably motivated by the evidence that regions of interest can provide better localization of individual entities and the ease to build their relational connections, the two-stage have become the de-facto approaches over a period of time. Typically, different approaches differ in how they represent the context. Mao \emph{et al.} \cite{mao2016generation} and Hu \emph{et al.} \cite{hu2016natural} use the whole image as a global context, while Yu \emph{et al.} \cite{yu2016modeling} directly pool visual feature from nearby objects as a way of modeling visual differences, showing that focus on the relationship between objects can achieve better results.
Furthermore, the context in \cite{nagaraja2016modeling, Hu2017} is regarded as weak supervision signals of unannotated objects, and multiple instances learning \cite{Dietterich1997} is then adopted to maximize the joint likelihood of all object pairs. However, the above modeling may oversimplify the number of contextual objects to a fixed size, \emph{e.g.}, one object as the contextual information. To this end, Zhang \emph{et al.} \cite{zhang2018grounding}  generates an attention map over all the objects as contextual information to approximate the combinatorial context configurations using a Variational Bayesian framework. For more detailed visual-language alignment, attention mechanisms are also widely adopted to fragment language to match the targets or contextual objects \cite{Deng2018, yu2018mattnet, zhuang2018parallel}. Unlike them, we think context can present arbitrarily within the whole scene and fully integrate them into a one-stage framework.

\noindent \textbf{One-Stage Visual Grounding.}
Before using one-stage object detectors, several methods have attempted to directly regress the bounding box from the whole image. However, these frameworks often suffer from a lower recall of objects, making them inferior to the two-stage counterparts. Some attention-based techniques are employed to enhance the local features of the target \cite{endo2017attention}. Besides, Yeh \emph{et al.} \cite{yeh2017interpretable} use a subwindow search to find the location that minimizes the energy function. Encouraged by the prominent one-stage detectors (\emph{e.g.}, YOLO \cite{Redmon2018}, SSD \cite{Liu2016}), many recent one-stage approaches have regarded proposals as grids in the feature map and directly regress the bounding box from the grid features responsible for detection \cite{yang2019fast, sadhu2019zero}. 
While achieving a large-margin improvement compared with that try to directly regress object from the entire image, such progress may attribute to the robust local features representation on the grid. Another line to improve one-stage visual grounding is to apply complex language modeling, such as decomposing the longer phrase into multiple parts \cite{yang2020improving}. In this work, we do not use complex techniques for language modeling. We show that by simply considering the context within the scene, our network can show competitive results.

\section{Landmark Feature Convolution}
We first summarize the most common convolutions, which we categorize into the family of \textit{point-based sampling} strategy, and along the way, discuss their relations, advantages and limitations. After that, we introduce our proposed \textit{region-based sampling} strategy, followed by the landmark feature convolution as well as its formulation and implementation.

\subsection{Point-based Sampling \label{sec:point-based}}

\begin{figure}[h]
\includegraphics[width=1\linewidth]{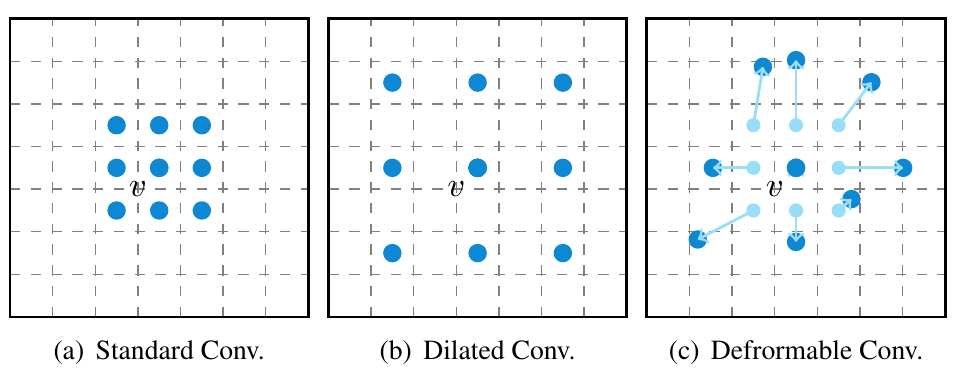}
\caption{A graphical view of different \textit{point-based} convolutions.}
\label{fig:convs}
\end{figure}

Given an input feature map $X=\{x_v:v\in V\}$  with node feature $x_v\in \mathbb{R}^c$, a \textit{point-based} convolution learns a representation vector $y_v$ by 
\begin{equation}
	y_v = \text{SUM}(\{W_{(u,v)} \cdot x_u: \forall u \in \mathcal{N}(v)\cup \{v\}\}),
	\label{eq:landmark}
\end{equation}
where $v \in V$ is the location of the node, $\mathcal{N}(v)$ is the neighborhood of node $v$ and $W_{(u,v)}$ parameterizes the spatial relation of node $u$ and node $v$. In the context of image feature maps, a node is the same as a location, so that we use both notations interchangeably. Different convolutions have different sampling strategies $\mathcal{N}(v)$. That is, how we sample nodes for convolution to represent the output vector $y_v$.

\noindent
\textbf{Standard Convolution.} In a $3 \times 3$ convolution kernel, a regurlar grid window $\mathcal{R}$ is used, which can be represented as a list of offests. Then the sampled neighbor $\mathcal{N}(v)$, or we called receptive field, is equal to $\{v+o:\forall o\in \mathcal{R}\}$, as shown in Figure \ref{fig:convs} (a).
Notably, the parameter $W$ is not shared among sampled locations, such that the spatial relations between node $v$ and its neighborhood nodes $u\in \mathcal{N}(v)$ can be explicitly captured. This property enables the convolution to detect meaningful patterns, such as line segments and corners. Theoretically, the receptive field grows as a convolutional layers stack, allowing deep CNNs to perform various high-level semantic tasks, such as object recognition, face detection, and semantic segmentation. However, the \textit{effecive} receptive field often occupies a fraction of the full \textit{theoratical} receptive field and converges to a Gaussian, making recognizing large objects and long-range modeling still challenge \cite{NIPS2016_c8067ad1}.

\noindent
\textbf{Dilated Convolution.}  To model long-range context, one solution is to increase the number of sampling points to enlarge receptive field, such as morphing the kernel size from $3\times3$ to $5\times5$. However, this substantially improves the number of parameters and brings the risk of over-fitting. To this end, the sampling window of a $3\times3$ kernel is dilated to a $5\times5$ grid window, resulting in a dilated convolution (in this case, dilation is 2) \cite{yu2015multi}, as shown in Figure \ref{fig:convs} (b). 

By enlarging the receptive field without introducing extra parameters, dilated convolution has become the de-facto technique to aggregate multi-scale context and hence has advanced a variety of researches \cite{chen2017deeplab,chen2017rethinking}. However, due to the sparse topology of sampling locations, dilated convolution can suffer from \textit{gridding} artifacts \cite{yu2017dilated,wang2018smoothed}. In visual grounding, this can hinder relation modeling of objects since their spatial positions can be arbitrary. 

\noindent
\textbf{Deformable Convolution.} Due to the fixed topology of sampling locations, the aforementioned CNNs are inherently limited to model large, unkown transformations \cite{dai2017deformable}. Deformable convolution relieves this issue by adding learnable 2D offsets to the regular grid via additional convolutional layers \cite{dai2017deformable}. That is, transforming $\mathcal{N}(v)=\{v+o:\forall o\in \mathcal{R}\}$ to $\mathcal{N}(v) = \{v+o+\Delta o:\forall o \in \mathcal{R}\}$, where $\Delta o$ is a leraned offset. After that, the node features are sampled from the transformed $\mathcal{N}(v)$ via bilinear interpolation. The illustrasion is shown in Figure \ref{fig:convs} (c). 

While deformable convolution excels in recognize objects by morphing the kernels to \textit{intra-object} geometries, little do we know that such deformation can generalize to model \textit{inter-object} relations, particularly in the context of visual grounding. One potential is that it may fail when modeling relations across very long distances since the learned offsets are expected to be constrained by the receptive field of their producers, \emph{i.e.} the standard CNNs.

\noindent
\textbf{Graph Convolution.} By regarding any pair of points have an edge, one can apply graph convolution on node $v$ to have a global receptive field. For example, Non-Local module \cite{wang2018non} update $y_v$ by 
\begin{equation}
y_v = \text{SUM}\{f(x_v, x_u)\cdot W \cdot x_u:\forall u \in V\}, 
\end{equation}
where $f(x_v, x_u)$ is a affinity between $x_v, x_u$, and $W$ is shared for all locations. Since $W$ is shared, the ability to represent spatial relations relies on $f(x_v, x_u)$, which requires $V$ has a suitable positional embedding. While applicable, how to effectively represent relative positional embedding still remains unclear.

\subsection{Region-based Sampling} 
\begin{figure}[h]
\includegraphics[width=1\textwidth]{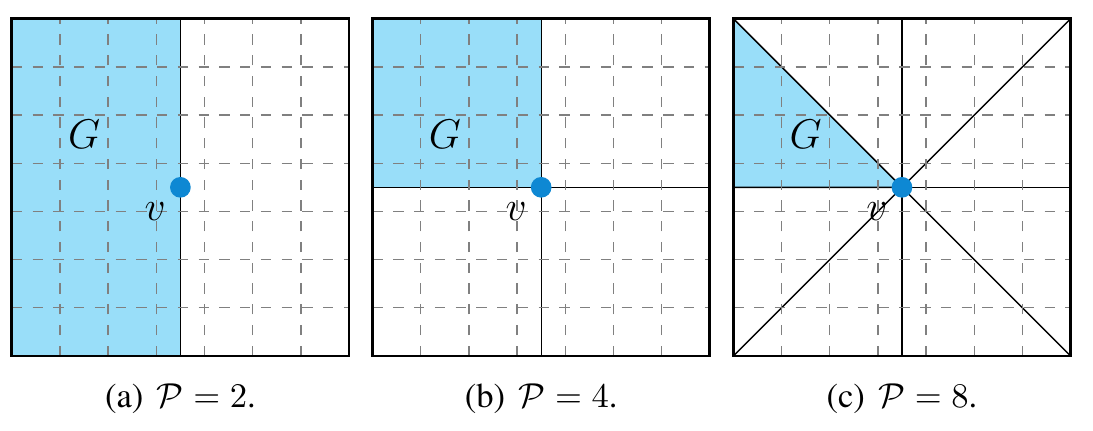}
\caption{Some variants of our \textit{region-based} convolution. $\mathcal{P}=k$ denotes that we divide the whole feature map $V$ into $k$ groups, \emph{i.e.} $\|\mathcal{P}(V)\|=k$. For clarity, only one group ($G$) is highlighted.}
\label{fig:region-based-sampling}
\end{figure}

To overcome the gridding artifacts and receptive field limitations of \textit{point-based} convolutions, we proposed a \textit{region-based} sampling strategy for convolution. That is, we set some axes on the node $v$ to part the whole feature map into several sub-regions and update the representation of $v$ by aggregating representations from each region. Formally, we update the representation of node $v$ by 
\begin{equation}
y_v = \text{SUM}(\{W_{(v, G)} \cdot h_G:\forall G\in \mathcal{P}_{v}(V)\}),
\end{equation}
where $\mathcal{P}_v$ denotes the partition function $\mathcal{P}$ over the input feature map $V$ based on node $v$, and $G$ is a group of nodes that shares similar spatial relation to node $v$, and $h_G$ is yielded from $G$, which we call \textit{landmark feature}. There are a variety of partitions $\mathcal{P}$, as shown in Figure \ref{fig:region-based-sampling}. For $\mathcal{P}=2$ in Figure \ref{fig:region-based-sampling} (a), the nodes are parted into two groups according to the vertical axis, such that one group is to the left of node $v$ and the other is to the right. By parameterizing two groups differently, the convolution can specialize in detecting horizontal spatial relations and hence help to ground the target. Namely, given the expression ``man to the right of car'', the likelihood of the nodes to the right of ``car'' will be raised.

To extract \textit{landmark feature} $h_G$, we apply a simple permutation invariant function, sharing the same spirit to those obtaining the entire graph's representation in graph classification \cite{kipf2016semi, xu2018how}. We use Max Pooling as the readout function, following the concept of landmark (namely, the most noticeable position).
To make the \textit{landmark feature} more descrimitive and spatial-aware, we can also use MLP or CNN to embed $h_G$. We find an additional one-layer MLP is sufficient. The $h_G$ is yielded as follow:
\begin{equation}
h_G = \text{MAX}(\{\text{ReLU}(W_G \cdot x_u):\forall u \in G\}), 
\end{equation}
where $W_G$ is the embedding parameter that is \textit{not} shared among different groups. Since $W_G$ is exclusive to each pariticular group, we do not need position embedding once choosing a suitable partition $\mathcal{P}$. In this paper, we empirically adopt $\mathcal{P}=4$ for modeling the most common relations (\emph{i.e.} ``left, right, on, below''), as shown in Figure \ref{fig:region-based-sampling} (b). Since our module update the representation of node $v$ with landmark features $\{h_G, G\in \mathcal{P}(V)\}$, we call it \textit{Landmark Feature Convolution}.

\noindent
\textbf{Implementation details}. We pay close attention to the efficiency of our proposed module. The biggest bottleneck is that we need to perform $k$ times Max Pooling to update $\{y_v:v\in V\}$, for $\mathcal{P} = k$. Noticing that \textit{landmark features} of adjacent nodes have overlapping sub-regions, computations can be reduced by dynamic programming. Assuming the input is the embedded feature map $\mathcal{X}_{G}$, we show how we compute $\mathcal{H}_{G}=\{h_{v}:\forall v \in V\}$ for the group highlighted in Figure \ref{fig:region-based-sampling} (b) with a few lines in Algorithm \ref{algo:dmp}, termed \textit{Dynamic Max Pooling}. Computing for other groups or partitions $\mathcal{P}$ can be implemented as straightforward. We also accelerate it with CUDA since each channel can run in parallel, which distinguishes our algorithm from those running RNNs over the feature map \cite{bell2016inside, NIPS2017_c22abfa3}. 

\begin{figure}[h]
\centering
\includegraphics[width=0.9\textwidth]{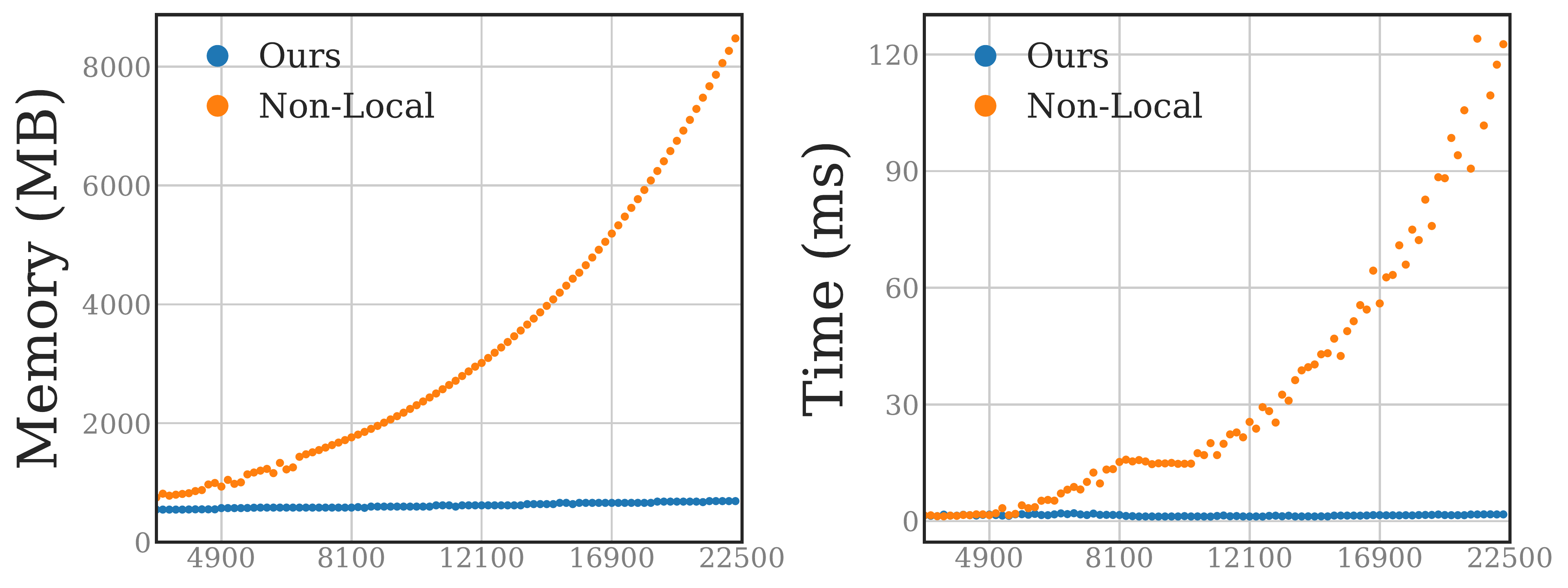}
\caption{Real time simulation of memory usage and running time. Different from Non-Local layer \cite{wang2018non}, our LFC is linear time-space complexity \emph{w.r.t.} $\|V\|$, and still enjoy a global receptive field.}
\label{fig:time-space}
\end{figure}

Overall, our algorithm has linear time-space complexity with respect to the number of nodes, \emph{i.e.} $\Theta(k\|V\|)$ where $k$ represents the number of partitions. Although having sequential operations, our implementation demonstrates its superiority over graph convolution layers, such as Non-Local layer \cite{wang2018non} or self-attention layer \cite{vaswani2017attention} whose time-space complexity is $\Theta(\|V\|^2)$, by simulations, as shown in Figure \ref{fig:time-space}. 

\begin{algorithm}[tb]
\small
\caption{Dynamic Max Pooling}
\begin{algorithmic}[1]
\REQUIRE An input $\mathcal{X} = \{x_{i,j}\}^{M\times N}$ where $x_{i,j} \in \mathbb{R}^c$.
\ENSURE An output $\mathcal{H} = \{h_{i,j}\}^{M\times N}$, where $h_{i,j} \in \mathbb{R}^{c}$.
\STATE $\mathcal{H} \leftarrow \mathcal{X}$
\FOR {$i\in [1, M]$}
\FOR{$j \in [1, N]$}
\STATE $h_{i,j} \leftarrow \text{MAX}(\{h_{i,j-1}, h_{i-1, j}\})$
\ENDFOR
\ENDFOR
\RETURN $\mathcal{H}$
\end{algorithmic}
\label{algo:dmp}
\end{algorithm}

\begin{figure*}
    \centering
    \includegraphics[width=1\linewidth]{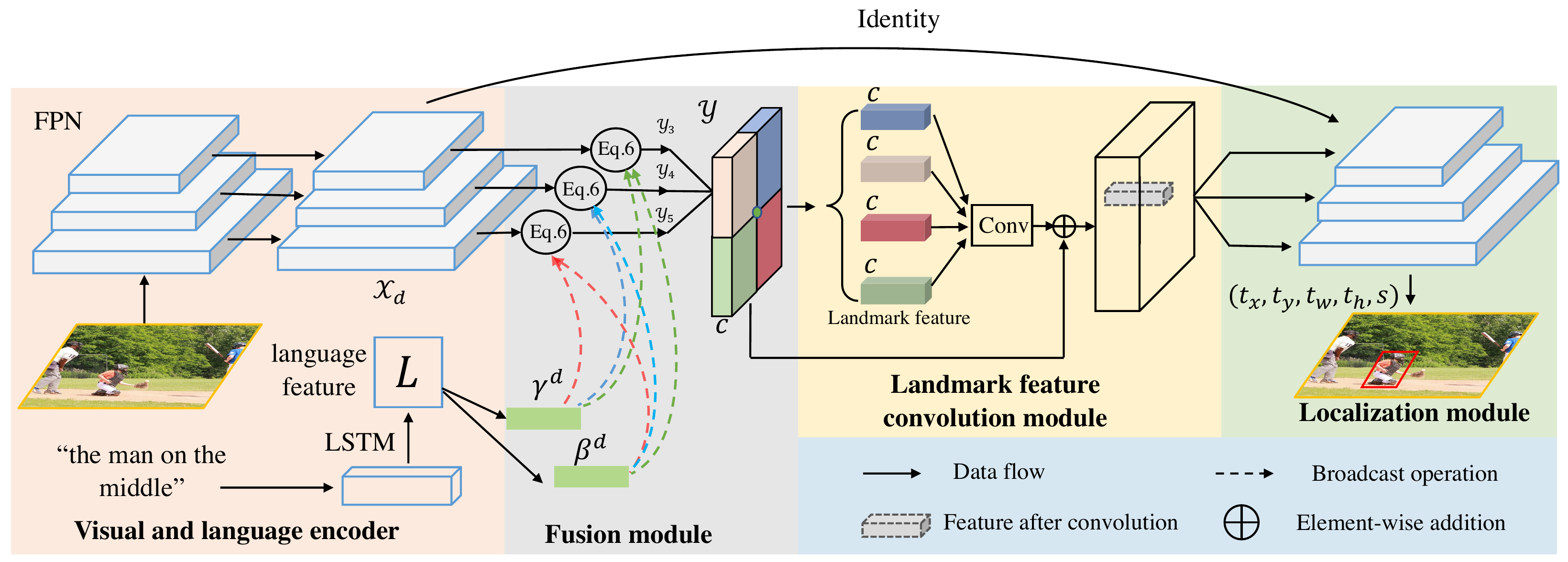}
    \caption{Our proposed LBYL-Net, which consists of four components: a visual and language encoder, a fusion module, a landmark feature convolution module and a localization module.}
    \label{fig:architecture}
\end{figure*}

\section{LBYL-Net}
Based on landmark feature convolution, we propose LBYL-Net. LBYL-Net consists of four components: a visual and language encoder, a fusion module, a landmark feature convolution module, and a localization module, which are introduced in the following, respectively.

\noindent \textbf{Visual and language encoder.} 
In Figure \ref{fig:architecture}, LBYL-Net firstly forwards the given image through a backbone network, where we use DarkNet-53 based Feature Pyramid Network (FPN) \cite{lin2017feature} to extract features from different scales. We choose the outputs from P3 to P5 levels of FPN as visual features $v \in \mathbb{R}^{c_d \times h_d \times w_d}$, where $d = {3, 4, 5}$ shows the $d$-th level. After that, we utilize a $1 \times 1$ convolution in $v$ to obtain feature maps with the same channel $c_v$ and concatenate the coordinate features with a 8 dimension position embedding vector, which is the same with prior work \cite{yang2019fast, yang2020improving}, such that we generate the fused feature maps $\mathcal{X}_d \in \mathbb{R}^{(c) \times h_d \times w_d}$, where $c=c_v+8$. 

For the language encoder, we firstly encode each word to dimension $c_l$ with a one-hot embedding given a language expression, and then a Bi-LSTM is applied to extract language feature $L \in \mathbb{R}^{c_l}$ to encode the whole expression. We also use BERT \cite{devlin2018bert} in place of LSTM to enhance language representation, following \cite{yang2019fast,yang2020improving}.
\par

\noindent \textbf{Fusion Module.} Given the generated language feature, we aim to obtain the maximum response of visual information conditioned on language. Therefore, we fuse visual and language features through a FiLM module \cite{perez2017film} and a $1\times 1$ convolution. FiLM applies an affine transformation in visual features $\mathcal{X}_d$ under the guidance of language $L$. The specific operations are as follows:

\begin{equation}
	\gamma^{d} = \text{MLP}^d_\gamma(L), \beta^{d} = \text{MLP}^d_\beta(L),
\end{equation}
and 
\begin{equation}
	\mathcal{Y}_d = \text{ReLU}(\text{Conv}(\text{ReLU}(\gamma^{d} \odot \mathcal{X}_d \oplus \beta^{d}))), 
\end{equation}
where $\text{MLP}^d_\gamma$ and $\text{MLP}^d_\beta$ are two one-layer MLP that maps language vector $L$ to coefficients $\gamma^{d}$ and $\beta^{d}$. Then we apply these coefficients to visual feature $\mathcal{X}_d$ from different FPN level followed by convolution and ReLU operations, yielding the output $\mathcal{Y}_d \in \mathbb{R}^{(c) \times h_d \times w_d}$, where $\odot$ and $ \oplus $ represent the broadcast element-wise multiplication and addition, respectively. After that, the feature of each position in $\mathcal{Y}_d$ might be adaptively responsible for different fine-grained properties, such as colors, positions, categories conditioned on language \cite{perez2017film}.

Prior to spatial relation modeling, we observe that FPN can hurt the performance. It could be that FPN distributes objects into different feature maps based on their scales, making modeling across-scale relation difficult. For example, given the relation ``painting over bed'', the ``painting'' hardly stands a chance of perceiving the ``bed'' if they are assigned to two separate feature maps. Simply summing the feature map to the intermediate size cures that problem, the same technique as BFPN \cite{pang2019libra}. In particular, we achieve this through max downsampling and bilinear upsampling, and finally:

\begin{equation}
	\mathcal{Y} = \frac{1}{3} \sum_{d=3}^{d=5} \mathcal{Y}_d.
\end{equation}

\begin{table*}[t]
    \centering
    \footnotesize
    \begin{tabular}{c|c|c|ccc|ccc|c|c}
    \toprule
    \multicolumn{1}{c|}{\multirow{2}{*}{Methods}} & \multicolumn{1}{c|}{\multirow{2}{*}{\makecell{Visual \\ Encoder}}} & \multicolumn{1}{c|}{\multirow{2}{*}{\makecell{Language \\ Encoder}}} & \multicolumn{3}{c|}{RefCOCO} & \multicolumn{3}{c|}{RefCOCO+} & \multicolumn{1}{c|}{RefCOCOg} & \multicolumn{1}{c}{\multirow{2}{*}{\makecell{Time\\(ms)}}}\\
    \cline{4-10}
 
    \multicolumn{1}{c|}{} & \multicolumn{1}{c|}{} & \multicolumn{1}{c|}{} & val & testA & \multicolumn{1}{c|}{testB} & val & testA & \multicolumn{1}{c|}{testB} & \multicolumn{1}{c|}{val}  
    \\ 
    \cline{1-11}
    \multicolumn{10}{c}{\textit{Two-stage methods}}\\
    \hline
    MMI \cite{mao2016generation}         	& VGG-16     	&-  	& - & 64.9 & 54.51 & - & 54.03 & 42.8 &  - & - \\
    Neg Bag \cite{nagaraja2016modeling}     & VGG-16     	&-		& - & 58.6 & 56.4  & - & -     & - & 49.5  & - \\
    CMN \cite{Hu2017}         				& VGG-16     	&LSTM	& - & 71.03 & 65.77 & - & 54.32 & 47.76 & - & - \\
    VC \cite{zhang2018grounding}          	& VGG-16     	&LSTM	& - & 73.33 & 67.44 & - & 50.86 & 58.03 & - \\
    ParallelAttn \cite{zhuang2018parallel} 	& VGG-16    	&LSTM	& - & 75.31 & 65.52 & - & 61.34 & 50.86 & - & - \\
    LGRAN \cite{wang2019neighbourhood}      & VGG-16     	&LSTM	& - & 76.6 & 66.4 &   - & 64.00 & 53.40 & 61.78 & - \\
    SLR \cite{yu2017joint}         			& ResNet-101    &LSTM	& 69.48 & 73.71 & 64.96 & 55.71 & 60.74 & 48.80 & - & -\\
    MAttNet \cite{yu2018mattnet}    		& ResNet-101    &LSTM   & 76.40 & 80.43 & 69.28 & 64.93 & 70.26 & 56.00 & - & 320 \\ 
    DGA \cite{yang2019dynamic} 				& ResNet-101    &LSTM   & - & 78.42 & 65.53 & - & 69.07 & 51.99 & - & 341\\
    CM-Att-Erase \cite{liu2019improving} 	& ResNet-101   	&LSTM   & \underline{78.35}	& \underline{83.14} & \underline{71.32} & \underline{68.09} & \underline{73.65} & \underline{58.03} & \underline{68.67} & -\\
    NMTree \cite{liu2019learning}  			& ResNet-101 	&TreeLSTM \cite{tai2015improved}& 76.41 &  81.21 & 70.09 & 66.46 & 72.02 & 57.52 & 64.62 & - \\
    \hline
    \hline
    \multicolumn{10}{c}{\textit{One-stage methods}}\\
    \hline
    RCCF \cite{liao2020real}	  						& DLA-34	  	&LSTM	& -       & 81.06	& 71.85	& -     & 70.35	& 56.32	& \textbf{65.73} & 25 \\
    YOLO-VG$^{\dagger}$ \cite{yang2019fast} 		& DarkNet-53 	&BERT	& 72.05	& 74.35	& 68.5 & 56.81 & 60.23 & 49.6 & 56.12 & 23 \\
    SQC-Base \cite{yang2020improving}  & DarkNet-53  	&BERT	& 76.59	& 78.22	& 73.25	& 63.23	& 66.64	& 55.53	& 60.96 & 26 \\
    SQC-Large \cite{yang2020improving} & DarkNet-53  	&BERT	& 77.63	& 80.45	& 72.3	& 63.59	& 68.36	& 56.81	& 63.12 & 36\\
    \hline
    Baseline \cite{yang2019fast} 						& DarkNet-53 	&LSTM	& 72.36& 73.86 & 65.93 & 57.98 & 63.97 &48.31 &47.25 & 24\\
    LBYL-Net {\tiny w/o LFC}         					& DarkNet-53  	&LSTM	& 77.43 &  80.75 & 70.68 & 64.84 &70.24 & 54.71 & 56.17 & 25 \\
    LBYL-Net      										& DarkNet-53  	&LSTM	& 78.76	& 82.18 & 71.91	& 66.67& 73.21 & 56.23	& 58.72 & 28\\
    LBYL-Net 											& DarkNet-53 	&BERT	& \textbf{79.67} & \textbf{82.91} & \textbf{74.15} & \textbf{68.64} & \textbf{73.38} & \textbf{59.49} & 62.70 & 30 \\
    \bottomrule
    \end{tabular}
    
    \caption{Performance comparisons on the RefCOCO, RefCOCO+, RefCOCOg. The best two-stage performance is highlighted with \underline{underline}, and the best one-stage performance is highlighted with \textbf{bold}.}

    \begin{minipage}[b]{1\columnwidth}
	\centering
    $\dagger$ indicates the result is adopted from \cite{yang2020improving}. 
	\end{minipage}
    \label{tab:result_coco}
\end{table*}

\begin{table}[h]
	\centering
	\scriptsize
	\begin{tabular}{c|c|c|c|c}
		\toprule
		Methods & \makecell{Visual\\Encoder} &  \makecell{Language\\Encoder} & \makecell{Pr@0.5 \\ (\%)} & \makecell[c]{Time\\(ms)} \\
		\hline
		\multicolumn{5}{c}{\textit{Two-stage methods}} \\
		\hline
		CMN\cite{Hu2017}      					&VGG-16  		&LSTM	& 28.33 & -\\
		VC \cite{zhang2018grounding}   			& VGG-16        &LSTM	& 31.13 & - \\
		Similarity Net \cite{wang2018learning}  & ResNet-101  	&-	& 34.54 & 184 \\
		CITE \cite{plummer2018conditional}   	& ResNet-101   	&-	& 35.07 & 196 \\
		MAttNet \cite{yu2018mattnet} 			& ResNet-101    &LSTM	& 29.04 & 320 \\
		DDPN$^{\ddagger}$ \cite{yu2018rethinking}    & ResNet-101&LSTM 	&63.00 & -\\
		\hline
		\hline
		\multicolumn{5}{c}{\textit{One-stage methods}} \\
		\hline
		ZSGNet 	\cite{sadhu2019zero}  		  	& ResNet-50 	&LSTM   & 58.63 & 25 \\
		RCCF 	\cite{liao2020real}    		  	& DLA-34 		&LSTM   & 63.79 & 25\\
		YOLO-VG \cite{yang2019fast} 			& DarkNet-53  	&LSTM	& 58.76 & 21 \\
		YOLO-VG \cite{yang2019fast} 			& DarkNet-53  	&BERT	& 59.30 & 38 \\
		SQC-Base \cite{yang2020improving} 		& DarkNet-53  	&BERT 	&64.33 & 26\\
		SQC-Large \cite{yang2020improving} 		& DarkNet-53  	&BERT 	&64.60 & 36\\
		\hline
		Baseline \cite{yang2019fast} 			& DarkNet-53 	&LSTM	& 59.28 & 24\\
		LBYL-Net {\tiny w/o LFC}  				& DarkNet-53 	&LSTM	& 62.59 & 25 \\
		LBYL-Net 		    					& DarkNet-53 	&LSTM	& 65.48 & 28\\
		LBYL-Net 								& DarkNet-53 	&BERT	&\textbf{67.47} & 30 \\
		\bottomrule
	\end{tabular}
	\begin{minipage}[b]{1\columnwidth}
	\centering
	\end{minipage}
	\caption{ Performance comparisons on the ReferitGame \cite{KazemzadehOrdonezMattenBergEMNLP14}.}
    \label{tab:referit}
\end{table}

\noindent \textbf{Landmark feature convolution module and localization module.}
To consider the landmark features, we apply a landmark feature convolution in feature map $\mathcal{Y}$, where we choose $\mathcal{P}=4$ (as shown in Figure \ref{fig:region-based-sampling} (b)). By DMP (\textit{dynamic max pooling}) and convolution, landmark features from four sub-regions are aggregated, yielding a direction-aware feature map. Afterward, we distribute the features to different FPN levels to account for the scale problem in general object detection. 

Finally, we feed them into the localization module, where we adopt an anchor-based box regression head in YOLOv3 as a detection head. The final output of LBYL-Net has a dimension of $KA \times h_d \times w_d$, where $A=3$ is the number of anchors and $K=5$ for $(t_x, t_y, t_w, t_h, s)$, where the first four values mean the bounding box offset relative to the pre-defined anchor and the last one is the confidence score indicating whether there is an object in this position. Following \cite{Redmon2018}, only the anchor with the largest IoU with the ground-truth bounding box is assigned as a positive sample; the rest are negative samples. Therefore, there is only one positive sample because we only want to find an object referred to by sentence. For the ranking loss, we maximize the distance between the positive sample and the negative samples. Thus a cross-entropy loss is employed, which can be viewed as MMI training defined in \cite{mao2016generation}. For the bounding box regression loss, we use an MSE loss to minimize the distance between the predicted bounding box and the ground-truth. The whole loss function consists of a localization term and a regression term:
\begin{equation}
\ell = \ell_{loc} + \beta \ell_{reg},
\label{eq: loss function}
\end{equation}
where $\beta$ is the hyper-parameter to balance two terms, and we empirically set $\beta=5$. The whole network is optimized with Adam \cite{Diederik2014} in an end-to-end manner.

\section{Experiments}
\label{sec: experiments}

\subsection{Implementation and Evaluation}
\noindent \textbf{Training.} A DarkNet-53 pre-trained on COCO is used as our backbone, and a cosine annealing strategy \cite{Ilya2016} is employed for optimization. We train our network with a learning rate $1e^{-4}$, weight decay $1e^{-4}$, batch size 64, with  GPUs. We do not use very high resolution for speed, although it could be helpful for the performance. The input images are resized $256 \times 256$ and employed on two GTX TITAN X. The total numbers of epochs are 100 for ReferitGame, RefCOCO, RefCOCO+ datasets, and 30 for the RefCOCOg dataset. 

The standard data augmentation methods in object detection are employed. We use random horizontal flip, random affine operations, and random color jitter. When horizontally flipping images, we need to flip the expressions simultaneously. \emph{e.g.}, replacing `left' with `right' and vice versa.

\noindent  \textbf{Evaluation.} We evaluate our method on ReferitGame \cite{KazemzadehOrdonezMattenBergEMNLP14}, RefCOCO \cite{yu2016modeling}, RefCOCO+ \cite{yu2016modeling} and RefCOCOg \cite{mao2016generation} visual grounding datasets. 
The evaluation metric is the same as that in \cite{lin2014microsoft}. Specifically, given a regressed bounding box of the referring object, we treat the regression as right if $\text{IoU} > 0.5$ between the ground-truth bounding box and prediction, termed as Pr@0.5. We also use Pr@0.75 for analyzing certain experiments.

\noindent \textbf{Settings.} We re-implement YOLO-VG \cite{yang2019fast} with an LSTM language encoder
as our baseline, which perform grounding on gird features individually, \emph{i.e.} only using $1\times1$ convolution in fusion module. 
We have small modifications to keep the same training scheme like ours, like learning rate and optimizer.  We see a slight improvement in accuracy in ReferitGame compared to the result reported in \cite{yang2019fast}. This will serve as our baseline for all of our experiments. We mainly report results with LSTM, unless specified.

\begin{table}[t]
    \footnotesize
    \centering
    \begin{tabular}{l|ll}
        \toprule
         Module & Pr@0.5(\%) & Pr@0.75(\%)  \\
        \hline
         \textit{Baseline} & 59.28 & 40.02 \\
         \hline
         \textit{+ FiLM} & \makecell[l]{60.99 (+1.71)} & 40.24 (+0.22) \\
         \hline
         \textit{\makecell[l]{+ FiLM \\ + BFPN}} & 62.59 (+1.71+1.60) & 41.00 (+0.22+0.76) \\
         \hline
         \textit{\makecell[l]{+ FiLM \\+ BFPN \\+ LFC}} & 65.48 (+1.71+1.60+2.87) & 44.31 (+0.22+0.76+3.31)\\
         \bottomrule
    \end{tabular}
    \caption{Ablation studies on ReferitGame. The numbers inside parentheses show the improvement upon the baseline.}
    \label{tab:ablation}
\end{table}

\subsection{Quantitative Results} 
\noindent \textbf{Comparisons with baselines.}
In summary, our LBYL-Net has about 6.2\%, 7.5\%, 8.6\%, 12.4\% absolute improvement on ReferitGame, RefCOCO, RefCOCO+, RefCOCOg, respectively, which demonstrates the effectiveness of our LBYL-Net. When a stronger language encoder is adopted, the performance can be further improved. The advance of our modification will be detailed in ablation studies.

\noindent  \textbf{Comparisons with state-of-the-art results.}
We compare our proposed LBYL-Net with state-of-the-art results of both one-stage and two-stage methods on ReferitGame, RefCOCO, RefCOCO+, RefCOCOg. The comparisons on ReferitGame are listed in Table \ref{tab:referit} and those on RefCOCO, RefCOCO+, RefCOCOg are listed in Table \ref{tab:result_coco}. Stronger visual and language representation can boost performance. For fair comparisons, we list the visual encoders and language encoders of these methods.

In ReferitGame, it is worth noting that the two-stage methods usually obtain poor results because they have no qualified proposals. We attribute the poor performance to the off-the-shelf detector that is not trained on ReferitGame. The evidence is that by using an end-to-end trainable RPN (region proposal network), the best result in two-stage methods can be achieved \cite{yu2018rethinking}. In COCO series datasets, \emph{e.g.}, RefCOCO, RefCOCO+, RefCOCOg, top results are usually achieved by two-stage methods since they adopt powerful detectors for COCO datasets. The detector helps to filter out irrelevant or noise regions prior to performing reasoning. However, our one-stage LBYL-Net still achieves competitive results among all the SOTA methods on RefCOCO, RefCOCO+, and the best result on ReferitGame. We show that not only is a one-stage pipeline advantageous to efficiency but can also achieve very strong performance by modeling long-range spatial relations. 

Another line to improve one-stage visual grounding is to better comprehend longer expression, especially for RefCOCOg, which contains more complex sentences. Although decomposing the expressions can achieve significant improvement \cite{liao2020real,yang2020improving}, we adopt a gloabl language representation for the sake of simplicity. On RefCOCOg, our model still improves the performance upon our baseline \cite{yang2019fast} by 12\% and 6\%, with LSTM and BERT, respectively, showing that modeling long-range spatial relations can help to comprehend longer sentences since these cases require more spatial relational cues to localize the target.

\begin{table}
    \footnotesize
    \centering
    \renewcommand\arraystretch{1.25}
    \begin{tabular}{c|cc|c}
        \toprule
         Ablation & \makecell{Pr@0.5(\%)} & \makecell{Pr@0.75(\%)} & \makecell[c]{Time \\ (ms)} \\
        \hline
         \textit{$\mathit{1\times1}$ Conv} & 62.59 & 41.00 &  25\\
         \hline
         \textit{Nonlocal NN} & 63.59 & 42.46 & 29\\
         \textit{Dilated Conv} & 63.85 & 42.90 & 26 \\
         \textit{Deform Conv} & 63.99 & 43.72 & 29 \\
         \textit{LFC} & \textbf{65.48} & \textbf{44.31} & 28 \\
         \bottomrule
    \end{tabular}
 \caption{Performance comparison to related convolution operations on ReferitGame.}
 \label{tab:LFC}
\end{table}

\begin{figure*}[h]
    \centering
    \includegraphics[width=0.96\columnwidth]{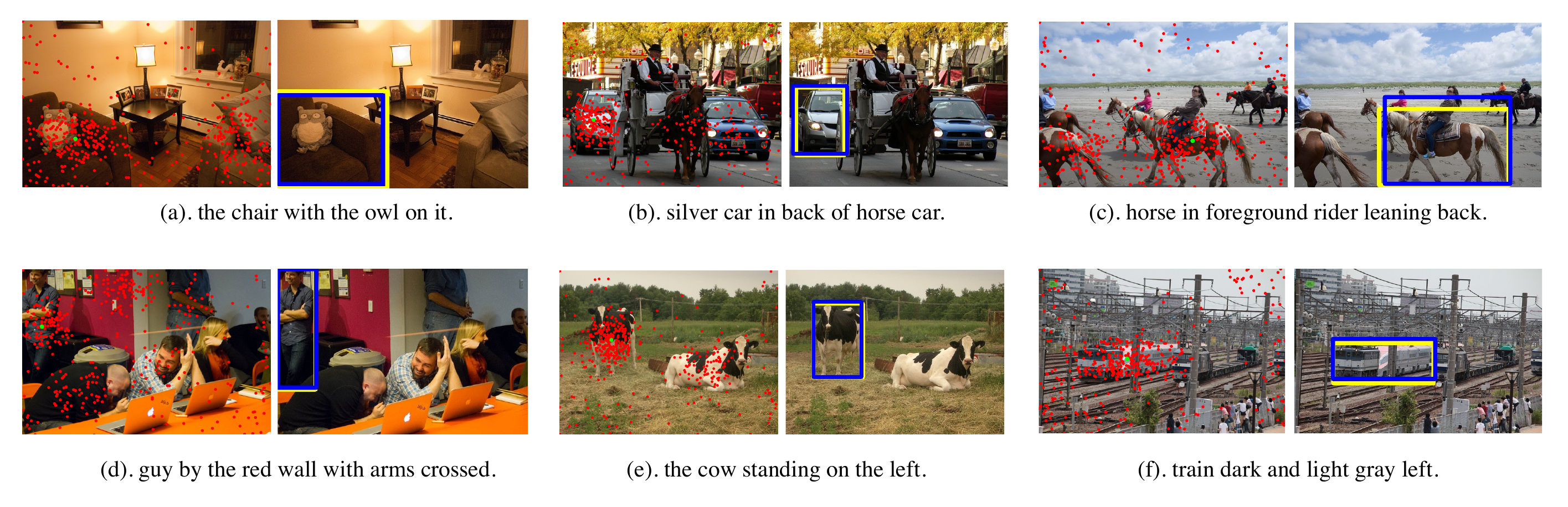}
    \caption{Visualizations of landmark positions and the grounding results. The images on the left side show the landmark positions (red dots) and the center of the predicted box (green dot). Notably, the predicted center receives information from landmark positions. The images on the right side show the ground-truths (yellow boxes) and the predictions (blue boxes).}
    \label{fig:visualize}
\end{figure*}

\subsection{Ablation Studies}
We conduct several ablation studies on ReferitGame \cite{KazemzadehOrdonezMattenBergEMNLP14} to reveal the effectiveness of our proposed LBYL-Net as well as our proposed \textit{Landmark Feature Convolution Module} (LFC). We additionally train three models upon this baseline by gradually replacing the `Concat-Conv' with FiLM \cite{perez2017film}, replacing the FPN with BFPN \cite{pang2019libra} and adding LFC, respectively. The results are shown in Table \ref{tab:ablation}. Thanks to the capacity of FiLM of fusing language and visual features, the performance is improved by 1.71\% under the metric Pr@0.5. The performance is further boosted by aligning visual features from different scales with BFPN by 1.6\%. However, the major improvement should be attributed to landmark feature convolution because it significantly raises the precision to 65.48\%. This can be more clearly validated under Pr@0.75. In this metric, LFC significantly improves the accuracy by more than 3\%, while the improvement of FiLM plus BFPN is marginally close to 1\%. 

\subsection{Effectiveness of LFC}
We first compare the performance of the LFC and the \text{point-based} convolutions discussed in Sec. \ref{sec:point-based}. We compare to Gaussian embedded Non-Local layer \cite{wang2018non}, Deformable convolution with kernel size 3 \cite{dai2017deformable}, and Dilated convolution with dilation 3. The result is shown in Table \ref{tab:LFC}. By comparing to $1\times1$ convolution, we show that a large receptive field is of central importance. We also see that Non-Local is inferior to other convolutions, probably due to its limitation of modeling relative spatial relation. With the merits of the global receptive field as well as spatial awareness, our LFC outperforms all of them. 

\begin{table}
\centering
\footnotesize
\begin{tabular}{c|c|cc|cc|c}
\toprule
Conv & \multicolumn{6}{c}{LFC} \\
\hline
w/o $\mathcal{P}$ & $\mathcal{P}=1$ & \multicolumn{2}{c|}{$\mathcal{P}=2$} & \multicolumn{2}{c|}{$\mathcal{P}=4$} & $\mathcal{P}=8$ \\
\hline
         \makecell[c]{\noindent\\\includegraphics[width=0.08\linewidth]{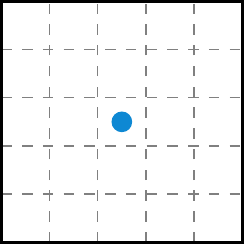}}& 
         \makecell[c]{\noindent\\\includegraphics[width=0.08\linewidth]{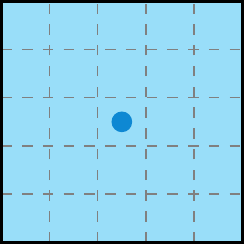}}
         & \makecell[c]{\noindent\\\includegraphics[width=0.08\linewidth]{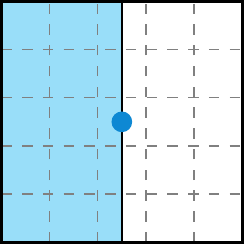}}
         & \makecell[c]{\noindent\\\includegraphics[width=0.08\linewidth]{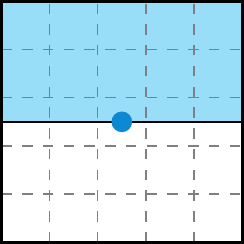}} 
         & \makecell[c]{\noindent\\\includegraphics[width=0.08\linewidth]{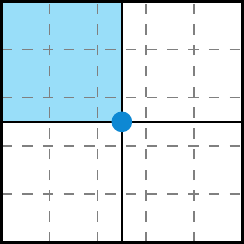}}
         & \makecell[c]{\noindent\\\includegraphics[width=0.08\linewidth]{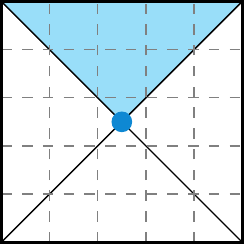}}
         & \makecell[c]{\noindent\\\includegraphics[width=0.08\linewidth]{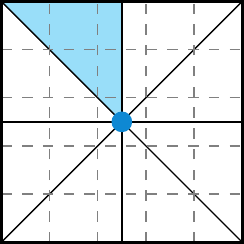}} \\
         \hline
62.59 & 63.90 & 65.05 &65.36& 65.48 & 65.26 & 65.58 \\
\bottomrule
\end{tabular}
\caption{Performance on ReferitGame (Pr@0.5) with different $\mathcal{P}$.}
\label{tab:partition}
\end{table}

\subsection{The Effect of Different Partitions}
We study the effect of various partitions, as shown in Table \ref{tab:partition}. When $\mathcal{P}=1$, our formulation of DMP (dynamic max pooling) degenerates to a global max pooling, yielding globally equal representations. We are surprised to find that such a simple global representation can boost performance. By considering spatial information, the performance can be further improved but reach saturation when $\mathcal{P}=4$. We hypothesis that this is dataset-related.

\subsection{Visualizations of Results and Landmarks}

Beyond effectiveness, the design of the \textit{landmark feature} $h \in \mathbb{R}^c$, which is max pooled from the sub-region, allows us to see where are focused over the whole feature map. In this way, we are able to take a small step toward interpretability in one-stage visual grounding. In particular, we can decode the landmark locations by \textit{argmax}. Since the \textit{landmark features} are pooled from a coarse feature map, to reflect on the original image size, we add a gaussian $\mathcal{G}(\mu, \sigma)$ for each landmark positions, where we choose $\mu=0, \sigma=1/3$. It is worth noting that there could be $c$ landmark positions since the dimension of $h$ is $c$. 

\par 
We visualize several examples of landmark features of the grounding centers, as well as the grounding results in Figure \ref{fig:visualize}. 
Many two-stage methods are typically motivated by the fact that the ROI-pooled features can provide better localization for individual objects and filter irrelevant background noise. We show that simply using Max-pooled features has a similar effect, \emph{i.e.} focusing on useful cues, but without resorting to extra supervision. In addition, while two-stage methods hold a strong assumption that contextual cues only come from a pre-defined set of objects, \emph{e.g.} 80 objects in COCO, we show that some of the cues outside of this distribution are also important, such as ``red wall'' as shown in Figure \ref{fig:visualize} (d). We return such a degree of freedom to the data itself.

\section{Conclusion}
In this work, we place emphasis on the relation modeling in one-stage visual grounding, and along this line of thought, propose a novel and simple LBYL-Net that shows competitive results over all state-of-the-art one-stage and two-stage methods. Central to our idea is to model long-range and spatial-aware features with \textit{Dynamic Max Pooling} (DMP) and \textit{Landmark Feature Convolution} (LFC), showing its superiority over related modules. We hope that our proposed LFC can also accelerate related researches, such as visual relation detection. 

\hspace*{\fill} \\
\noindent \textbf{Acknowledgement.} The work was supported by National Key R\&D Program of China (2018AAA0100704), 
NSFC \#61932020, Science and Technology Commission of Shanghai Municipality (Grant No. 20ZR1436000), and ``Shuguang Program'' supported by Shanghai Education Development Foundation and Shanghai Municipal Education Commission.

\clearpage
{\small
\bibliographystyle{ieee_fullname}
\bibliography{egbib}
}
\end{document}